\documentclass[twocolumn]{article}

\usepackage[english]{babel}
\usepackage{csquotes}

\usepackage[
backend=biber,
style=numeric,
]{biblatex}

\usepackage[letterpaper,top=2cm,bottom=2cm,left=3cm,right=3cm,marginparwidth=1.75cm]{geometry}

\usepackage{amsmath}
\usepackage{graphicx}
\usepackage[colorlinks=true, allcolors=blue]{hyperref}

\title{Violation of Expectation via Metacognitive Prompting Reduces Theory of Mind Prediction Error in Large Language Models}

\author{Courtland Leer, Vincent Trost, Vineeth Voruganti\\[1ex]
  \textit{Plastic Labs Inc.} \\
  \texttt{[courtland, vince, vineeth]@plasticlabs.ai}
}

\begin{document}
\maketitle

\begin{abstract}
Recent research shows that Large Language Models (LLMs) exhibit a compelling level of proficiency in Theory of Mind (ToM) tasks. This ability to impute unobservable mental states to others is vital to human social cognition and may prove equally important in principal-agent relations between individual humans and Artificial Intelligences (AIs). In this paper, we explore how a mechanism studied in developmental psychology known as Violation of Expectation (VoE) can be implemented to reduce errors in LLM prediction about users by leveraging emergent ToM affordances. And we introduce a \textit{metacognitive prompting} framework to apply VoE in the context of an AI tutor. By storing and retrieving facts derived in cases where LLM expectation about the user was violated, we find that LLMs are able to learn about users in ways that echo theories of human learning. Finally, we discuss latent hazards and augmentative opportunities associated with modeling user psychology and propose ways to mitigate risk along with possible directions for future inquiry.
\end{abstract}

\section{Motivation}
Plastic Labs is a research-driven product company whose mission is to eliminate the principal-agent problem \cite{jensen2019theory} horizontally across human-AI interaction. In a near future of abundant intelligence, every human becomes a potent principal and every service an agentic AI. Alignment of incentives and information, then, must occur at the scale of the individual. Enabling models to deeply understand and cohere to user psychology will be critical and underscores the importance of research at the intersection of human and machine learning.

\section{Introduction}
Large Language Models (LLMs) have been shown to have a number of emergent abilities \cite{wei2022emergent}. Among those is Theory of Mind (ToM), defined as ``the ability to impute unobservable mental states to others'' \cite{kosinski2023theory}. The emergence of this specific capability is of significant interest, as it promises LLMs with the ability to empathize and develop strong psychological models of others, as humans do naturally.

But how do you best position LLMs to demonstrate these qualities? Typical methods posit that connecting data sources deemed personal (e.g.\ email, documents, notes, activity, etc.) is sufficient for learning about a user. Yet these methods assume individual persons are merely the aggregate of their intentionally produced, often superficial, digital artifacts. Critical context is lacking — the kind of psychological data humans automatically glean from social cognition and use in ToM (e.g.\ beliefs, emotions, desires, thoughts, intentions, knowledge, history, etc.). 

We propose an entirely passive approach to collect this data, informed by how developmental psychology suggests humans begin constructing models of the world from the earliest stages \cite{onishi200515}. This cognitive mechanism, known as Violation of Expectation (VoE) \cite{brod2022explicitly}, compares predictions about environments against sense data from experience to learn from the difference, i.e.\ errors in prediction. 

Inspired by prompting methodologies like Chain-of-Thought \cite{wei2022chain} and Metaprompt Programming \cite{reynolds2021prompt}, we design a \textit{metacognitive prompting} framework for LLMs to mimic the VoE learning process. And we show that VoE-data-informed social reasoning about users results in less ToM prediction error.

This paper has the following two objectives:
\begin{enumerate}
    \item Demonstrate the general utility of a metacognitve prompting framework for VoE in reducing ToM prediction error in a domain-specific application —  \href{https://chat.bloombot.ai}{Bloom}, a free AI tutor available on the web and via Discord.
    \item Discuss at length opportunities for future work, including the practical and philosophical implications of this emergent capability to create psychological renderings of humans and ways to leverage confidential computing environments to secure them.
\end{enumerate}

We use OpenAI's GPT-4\footnote{GPT-4 32k version: 0613} API in the entirety of this experiment and its evaluation.
\begin{figure*}
\centering
\includegraphics[width=\textwidth]{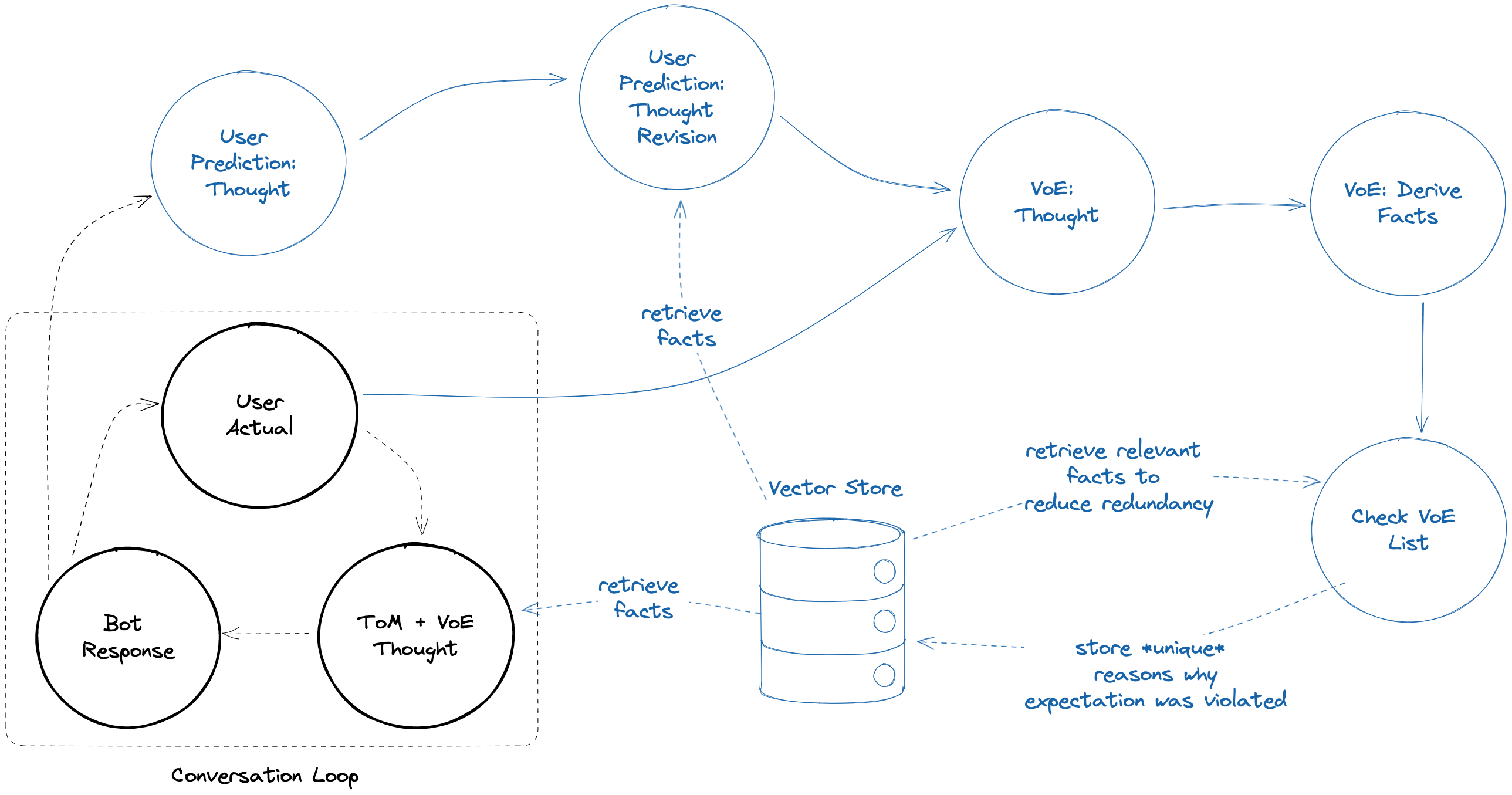}
\caption{\label{fig:diagram}Framework. Contained in the grey dotted box is an application's core conversation loop (e.g.\ our AI tutor, \href{https://chat.bloombot.ai}{Bloom}) and drawn in blue is the metacognitive prompting framework described in section \ref{sec:methods}.}
\end{figure*}
\section{Framing and Related Work}
\textbf{Predictive Coding and Theory of Mind}.
While not yet a complete theory, Predictive Coding (PC) continues to gain traction as a framework for understanding how modeling and learning occur in biological brains. At a high level, PC hypothesizes that mental models of reality are built and employed by comparing predictions about environments with sensory perception \cite{schultz1997neural}. PC-inspired approaches to machine learning show great initial promise as biologically plausible AI training methodologies \cite{salvatori2023braininspired}.

ToM is the ability of some organisms to, despite lacking direct access to any experience but their own, ascribe mental states to others. Notably, PC ``may provide an important new window on the neural computations underlying theory of mind'' as ToM ``exhibit[s] a key signature of predictive coding: reduced activity to predictable stimuli'' \cite{koster2013theory}. That is, when others behave in line with our predictions (i.e.\ our ToM projections are accurate) less is learned. And the inverse applies — the prediction errors enhance our capacity for high-fidelity ToM over time.

\textbf{Emergent Behaviors}.
Researchers have long been interested in getting large language models to exhibit ``thinking'' and ``reasoning'' behaviors. A number of papers have been influential in pioneering ways to elicit these via prompting \cite{brown2020language, wei2022chain, kojima2023large, yao2023react}. As model architectures have scaled, these abilities appear to have emerged without explicit training \cite{wei2022emergent}. While there's considerable debate concerning the distinction between ``emergent abilities'' and ``in-context learning,'' \cite{lu2023emergent} these phenomena display clear utility, regardless of taxonomy.

Quantifying just how vast the space of latent ``overhung'' LLM capabilities really is constitutes a major area of formal and enthusiast-driven inquiry. ToM is one such highly compelling research domain. Kocinski \cite{kosinski2023theory} shows that the OpenAI GPT-series of models possess the ability to pass fundamental developmental behavior tests. Some papers demonstrate how to improve these abilities \cite{moghaddam2023boosting} and others analyze these methods critically, questioning the premise of ToM emerging in LLMs \cite{ullman2023large, shapira2023clever}. 

Adjacently, there's a clear trend of researchers pushing the limit of what types of cognitive tasks can be offloaded to LLMs. In order to scale supervision, eliminate human feedback, avoid evasive responses, and have transparent governing principles, Anthropic has experimented with delegating the work of human feedback to the LLM itself in their ``constitutional'' approach  \cite{bai2022constitutional}. Other papers looking to achieve similar types of outcomes, without needing to update model weights, rely on in-context methods entirely \cite{shinn2023reflexion, zhou2023solving}.

\textbf{Violation of Expectation}.
One prime task candidate, which leverages emergent ToM abilities, is VoE. Similar to explanations from PC theories of cognition, VoE is an explicit mechanism that reduces prediction errors to learn about reality. 

While much of VoE happens in the unconscious mind and from an early age \cite{onishi200515}, research suggests that deliberate prediction making and error reduction also leads to enhanced learning outcomes \cite{brod2022explicitly}. 

Just as PC may play a role in ToM, VoE is a lightweight framework for identifying the data needed to minimize ToM error. Predicts are generated, compared against percepts, and learning is derived from the difference.

\textbf{Prompting Paradigms}.
\label{sec:prompting}
Chain-of-Thought \cite{wei2022chain} prompting clearly shows that LLMs are capable ``reasoning'' generators and that this species of prompting can reduce the probability of generating incorrect answers. Yet, as this method is limited to one inference, the model often disregards that reasoning, especially during ToM-related tasks.

Metaprompt Programming \cite{reynolds2021prompt} seeks to solve the laborious process of manually generating task-specific prompts (which are more efficacious than general ones) by leveraging LLMs' ability to few-shot prompt themselves dynamically.

Deliberate VoE as learning method, ToM, and these prompting approaches all echo the human phenomenon of metacognition — put simply, thinking about thought. In the next section we introduce a \textit{metacognitive prompting} framework in which the LLM generates ToM ``thoughts'' to be used in further generation as part of a VoE framework to passively acquire psychological data about the user.

\section{Methods}
\label{sec:methods}
The cognitive mechanism VoE can be broken down into two circular steps: 
\begin{enumerate}
\item Making predictions about reality based on past learning.
\item Learning from the delta between predictions and reality. 
\end{enumerate}
In the typical chat setting of a conversational LLM application, this means making a prediction about the next user input and comparing that with the actual input in order to derive psychological facts about the user at each conversational turn. We employ metacognitive prompting across both core parts of our framework shown in Figure \ref{fig:diagram}: our \textit{user prediction task} and our \textit{violation of expectation task}.
\begin{figure*}[ht]
\centering
\begin{tabular}{|l|r|r|r|r|}
\hline
\textbf{Assessment} & \textbf{VoE N} & \textbf{VoE Pct} & \textbf{Non-VoE N} & \textbf{Non-VoE Pct} \\
\hline
1. Very & 35 & 0.106 & 96 & 0.151 \\
2. Somewhat & 78 & 0.237 & 77 & 0.121 \\
3. Neutral & 17 & 0.052 & 22 & 0.035 \\
4. Poorly & 90 & 0.274 & 170 & 0.267 \\
5. Wrong & 109 & 0.331 & 272 & 0.427 \\
\hline
\end{tabular}
\caption{Results from A/B test in the  \href{https://chat.bloombot.ai}{Bloom} Web UI.}
\label{fig:results}
\end{figure*}

\subsection{Metacognitive Prompting}
Synthesized from the influences mentioned in Section \ref{sec:prompting}, we introduce the concept of \textit{metacognitive prompting}. The core idea is prompting the model to generate ``thoughts'' about an assigned task, then using those ``thoughts'' as useful context in the following inference steps. We find that in practice, this method of forced metacogntion enhances LLM ability to take context into account for ToM tasks (more discussion in Section \ref{sec:measuring_coherence}, ``Measuring Coherence'').

\textbf{Task 1: User Prediction and Revision}.
Given history of the current conversation, we prompt the LLM to generate a ToM thought including:
\begin{itemize}
    \item Reasoning about the user's internal mental state
    \item Likely possibilities for the next user input
    \item A list of any additional data that would be useful to improve the prediction
\end{itemize}
The list serves as a query over a vector store to retrieve relevant VoE derived user facts from prior interactions.

We then prompt the model in a separate inference to revise the original ToM thought given new information, i.e.\ the retrieved facts that have been derived and stored by VoE. These facts are psychological in nature and taken into account to produce a revision with reduced prediction error.

\textbf{Task 2: Violation of Expectation and Revision}.
We employ the same prompting paradigm again in the VoE implementation. 

The first step is to generate a ``thought''  about the difference between prediction and reality in the previous user prediction task. This compares \textit{expectation} — the revised user prediction — with \textit{violation} — the actual user input. That is, how was expectation violated? If there were errors in the user predictions, what were they and why?
 
This thought is sent to the next step, which generates a fact (or list of facts). In this step, we include the following:
\begin{itemize}
\item Most recent LLM message sent to the user
\item Revised user prediction thought
\item Actual user response
\item Thought about how expectation was violated
\end{itemize}
Given this context, fact(s) relevant to the user's actual response are generated. This generation constitutes what was learned from VoE, i.e.\ prediction errors in ToM. 

Finally, we run a simple redundancy check on the derived facts, then write them to a vector store. We used the OpenAI Embeddings API for the experiment in this paper.

\section{Experiments}
\label{sec:experiment}
Our experiment aims to show that using VoE derived data reduces error in LLM prediction about the next user input. This is especially useful and testable in conversations, so we use data from our AI tutor,  Bloom, which is specifically prompted to keep a conversation moving forward to produce learning outcomes for users.

Traditional conversation datasets often lean toward trivial dialogue, while instruction-following datasets are predominantly one-sided and transactional. Such datasets lack interpersonal dynamics, offering limited scope for substantive social cognition. Thus, our experiment employs an A/B test with two versions of our AI tutor, conversations with which more closely reflect psychologically-informative social interactions between humans.

The first version — the control — relies solely on past conversation to predict what the user will say next. Yet the second version — the experimental — uses our metacognitive prompting framework in the background to make predictions. Crucially, and as described in Section \ref{sec:methods}, the framework leverages VoE to increase the amount of information at the model's disposal to predict user responses. These VoE facts are introduced to the AI tutor through the additional ``thought revision'' phase in the conversational loop, allowing it to reduce prediction error and psychologically cohere itself more closely to the user. 

We use the same LLM — GPT-4 — to classify how well each version predicts each user input. Its assessment is useful to discern whether VoE data can reduce LLM prediction error as LLMs are competent arbiters of token similarity.

We do so by prompting GPT-4 to choose from 5 options that assess the degree to which a generated user prediction thought is accurate. The choices include ``very,'' ``somewhat,'' ``neutral,'' ``poorly,'' and ``wrong.'' We include the most recent AI message, thought prediction, and actual user response in the context window. The evaluation scripts can be found on GitHub\footnote{https://github.com/plastic-labs/voe-paper-eval}.

\section{Results}
\textbf{Dataset}.
\label{sec:dataset}
This experiment uses a dataset of conversations users had with Bloom. We built it by running an A/B test on the backend of Bloom's web interface. Only conversations of 3 or more turns are included. We recorded 59 conversations where the VoE version was active and 55 conversations where it was not. Within those, we collected 329 message examples from the VoE version and 637 from the non-VoE version. More on that difference in the ``Considerations'' paragraph in this section.

\textbf{Chi Square Test}.
We chose to give the model freedom to choose more granular assessments like values ``somewhat'', ``neutral'', and ``poorly'' rather than forcing it into a binary classification, but we found it barely used the ``neutral'' option. On a five-point scale, the top two ratings (``very'' and ``somewhat'' predictions) are grouped as ``good'', neutral ratings are omitted from the analysis, and the lowest two ratings (``poorly'' and ``wrong'') are grouped as ``bad''. 

We want to test the independence of two categorical variables: \textit{assessment} (good or bad) and \textit{group} (VoE or non-VoE). The observed frequencies are given in the following table:

\begin{center}
\begin{tabular}{|c|c|c|}
\hline
 & VoE & Non-VoE \\
\hline
Good & 113 & 173 \\
Bad & 199 & 442 \\
\hline
\end{tabular}
\end{center}

The Chi-square test statistic is calculated as:

\[
\chi^2 = \sum \frac{(O_{ij} - E_{ij})^2}{E_{ij}}
\]

where $O_{ij}$ are the observed frequencies and $E_{ij}$ are the expected frequencies under the null hypothesis of independence. The expected frequencies are calculated as:

\[
E_{ij} = \frac{(row\ total_i)(column\ total_j)}{grand\ total}
\]

For each cell, we calculate the expected frequency and then the contribution to the Chi-square statistic. The degrees of freedom for the test are $(R - 1)(C - 1)$, where $R$ is the number of rows and $C$ is the number of columns. 

The Chi-Square Test indicated a significant relationship between assessment and group, $X^2 (1, 927) = 5.97$, $p < .05$, such that VoE predictions were evaluated as good more often than expected and bad less often than expected. These results support our hypothesis that augmenting the Bloom chatbot with VoE reasoning reduces the model's error in predicting user inputs.

\textbf{Reducing Prediction Errors}.
The VoE version showed a significant reduction in prediction errors, resulting in fewer ``wrong'' values being generated. Overall, the VoE version exhibited a smoothing effect, enhancing the consistency of predictions. Although there was a slight decrease in ``very'' predictions, a relative increase of 51\% in ``somewhat'' values was observed. This shift suggests an improvement in prediction fidelity, balancing out extreme predictions with more moderate ones. Notably, the VoE version generated 22.4\% fewer ``wrong'' predictions compared to the Non-VoE version.

\begin{figure}
\centering
\includegraphics[width=0.4\textwidth]{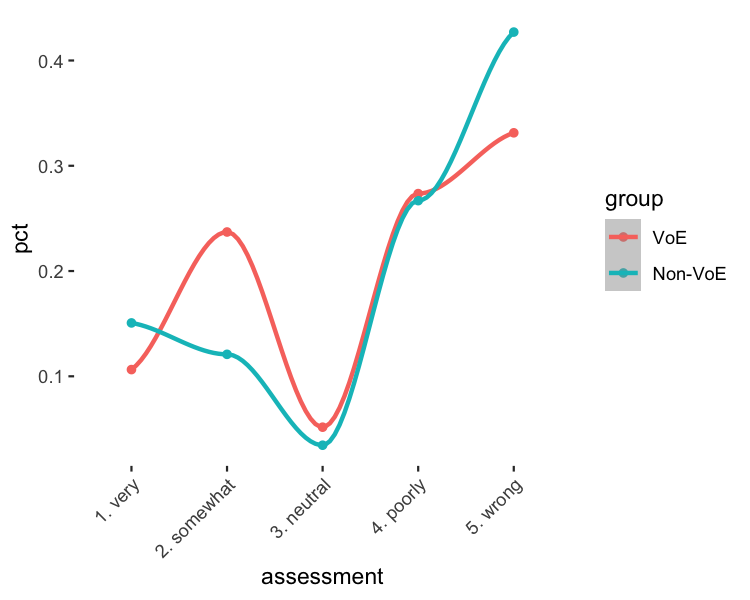}
\caption{\label{fig:ggplot-line}Plot of results found in Figure \ref{fig:results}. VoE smooths the distribution of predictions, reducing prediction error by learning from prior generations. This echoes accounts of human learning described in PC and VoE theories.}
\end{figure}
\textbf{Considerations}.
\label{sec:considerations}
The inherent nature of VoE is to improve and refine over time. As the vector store becomes populated with more data, the accuracy and relevance of VoE's outputs are expected to increase, enabling more valuable responses for users. 

It's important to note the presence of latency in VoE Bloom. This likely contributed to the reduction in conversation turns to nearly half that of the non-VoE Bloom. Nevertheless, the fact we observe a statistical difference between the groups given this discrepancy in data size is noteworthy.

There are a number of other practical factors in our data that might inhibit our ability to accurately measure the degree to which user prediction error was minimized. We used our conversational AI tutor's data for this study, which is subject to various issues that are being faced by all consumer-facing AI applications. This technology is new, and people are still learning how to interface with it. Many users ask Bloom to search the internet, do mathematical computations, or other things that aren't well served by the prompting framework around GPT-4.

Finally, it's of conceptual interest that LLMs can, from prompting alone, reduce prediction errors via mechanisms similar to those posited by PC and VoE theories of human cognition.

\section{Future Work and Beyond}
\subsection{Improvements}
\textbf{Retrieval Augmented Generation}.
Currently, our VoE fact retrieval schemes are quite naive. The ``thought'' generation steps are prompted to generate thoughts \textit{and} additional data points that would help improve the prediction. Those additional data points serve as a basic semantic similarity query over a vector store of OpenAI embeddings, and we select top \textit{k} entries. Much could be done to improve this workflow, from training custom embedding models to improving the retrieval method. We also draw inspiration from the FLARE paper \cite{jiang2023active} and note the improved generation results that come from forecasting a conversation and incorporating that into the context window.

\textbf{Training/Fine-Tuning}.
Similar to how instruction tuning yielded much improved results in decoder-only LLMs, we believe that ToM tuning is a task that could yield better psychological models. The task of following instructions is a sufficiently abstract idea. Making ToM predictions falls into the same category. 

\subsection{Evaluation}
\textbf{Assessing Theory of Mind}.
The authors of ``Clever Hans or Neural Theory of Mind? Stress Testing Social Reasoning in Large Language Models'' \cite{shapira2023clever} explicitly state that ``the consequences of the success of these tests do not straightforwardly transfer from humans to models'' and speak at length to the evolving landscape of datasets and evaluation methods aimed at machines instead of humans. The debate about whether or not LLMs ``have'' ToM is likely to continue and more semantic definitional work also needs to be done, but what's undeniable is the utility of this capability. Specifically interesting is boosting the performance of LLMs to minimize user prediction error, as much may become possible as a result of gains in that domain.

\textbf{Measuring Coherence}.
\label{sec:measuring_coherence}
For this paper, we exclusively leverage OpenAI's closed-source models behind their API endpoints. Because of this, we are fundamentally limited in the ways in which we can measure user prediction error. In order to remain consistent, we have the same LLM that is generating the ToM predictions generate a naive assessment of its accuracy, which is described more in Section \ref{sec:experiment}.

Experiments with open source LLMs allow much more granular evaluation. E.g.\ computing the conditional loss over a sequence of tokens or creating new datasets by employing human labelers to train an evaluation model. Establishing a more rigorous standard around evaluating ToM predictions with multi-turn interpersonal conversation data is an imperative area of work as well. 

The space of open source models is relatively untested in regard to ToM abilities. Comprehensive study of how the open source model stable performs on already existing tasks is a crucial next step.

Still further challenges exist in establishing reliable evaluation methods for measuring LLM coherence to users. Each user possesses not only unique psychological properties, but varying levels of awareness of that psychological profile. These subjective limitations demand novel approaches, research into which is only now becoming possible.

\subsection{Utility}
\textbf{Infrastructure}.
In a world of abundant synthetic intelligence, if vertical-specific AI applications remain viable, they will seek to outperform foundational models within their narrow purview. Redundantly solving personalization and psychological modeling problems represents unnecessary development and data governance overhead \textit{and} risks contaminating datasets. Nor is it in the security or temporal interest of users to share such data. Horizontal frameworks and protocols are needed to safely and efficiently manage these data flows, improve user experience, and align incentives.

\textbf{Products}.
Ability to robustly model user psychology and make ToM predictions about internal mental states represents novel opportunity for the frontier of goods and services. Bespoke multi-modal content generation, high-fidelity human social simulation, on-demand disposable software, atomization of services, instant personalization, and more could all become possible. Much work will be needed to explore this design space.

\subsection{Security}
While ToM data holds powerful personalization potential, the management and use of that data entails profound responsibility and promises significant hazards. Such data, rich with insights into internal user identity and future behavior suggests immense utility. Yet, this utility makes it a likely target for misuse or object of mishandling — more so given the remarkable inferential capabilities of LLMs.

Security implications are far-reaching, from privacy invasion and identity theft to manipulation and discrimination. Moreover, any breach of trust impacts not just individual users, but the reputation and success of organizations employing it. Below is a non-exhaustive list of future work needed to secure such data throughout its lifecycle. 

\textbf{Encryption and Custody}.
Due to the sensitive, individual nature of ToM data, encryption is a bare minimum security requirement, and there are strong arguments to be made for direct user key ownership. Formal investigations into appropriate solutions to both are needed.

The process of transforming plaintext to ciphertext safeguards the data from keyless access. Several methods of encryption, including symmetric methods like the Advanced Encryption Standard, which uses the same key for encryption and decryption, and asymmetric encryption methods like RSA, which uses two keys, a public key for encryption and a private key for decryption \cite{rivest1978method}, are plausible candidates.

Models for key management will dictate the exact implementation of encryption against the data. A method such as Shamir's secret sharing can be used to split the decryption key between a user and a trusted platform hosting the data \cite{dawson1994breadth}. However, the intimate nature of the data may still warrant user ownership, preventing even the platform from accessing the data. 

\textbf{Confidential Computing}.
This relatively new technology encrypts data in use (i.e. during processing). Confidential computing is a step beyond traditional methods that encrypt data at rest and in transit, thus providing a more comprehensive data protection framework. It leverages hardware-based Trusted Execution Environments (TEEs) to protect data during computation, enabling sensitive data to be processed in the cloud or third-party environments without exposing it to the rest of the system \cite{confidential2020confidential}.

Further work can determine architectures for safely mounting user data into TEEs, decrypting, and then using it to improve interactions between users and LLMs. Work to explore how to create a scalable and performant design that does not sacrifice security is needed. Additional considerations need to be made for securely using data with third-party LLM APIs such as OpenAI's GPT-4 as opposed to self-hosted models.

\textbf{Policy-Based Access Control}.
Policy-Based Access Control (or Attribute Based Policy Control) is a method used to regulate who or what can view or use resources in a computing environment \cite{hu2013guide}. It's based on creating, managing, and enforcing rules for accessing resources to define the conditions under which access is granted or denied.

Policies that can be applied on the data to ensure principles of least privilege to client applications and prevent data leakage are directions for further inquiry. LLM applications could be used to extend the policies to allow attributes based on the content of the data, such as grouping by topic.

\textbf{Frontier Security}.
LLMs' powerful inference abilities place them in a new category of digital actors. New paradigms of protection and security will be required. LLMs themselves might be leveraged to proactively monitor and obfuscate user activity or destroy unwanted statistical relationships. The advent of instant personalization may even make persistent application-side user accounts irrelevant or unsustainably hazardous.

\subsection{Philosophy}
\textbf{Extended Self}.
Chalmers and Clark argued in 1998 that minds can be said to extend into the physical world and still legitimately be considered part of personal cognition \cite{Clark1998-CLATEM}. High-fidelity human psychological renderings in AI agents suggest the potential for human agency and identity to extend in similar ways. Unanswered legal, metaphysical, and ethical questions arise from this prospect. 

\textbf{Phenomenology}.
When humans impute mental states to others, presumably that assignment is grounded in lived personal experience. That is, we can imagine other people having experiences because we have had similar experiences ourselves. Additionally, we share with the objects of our ToM a genetic schema and physical substrate for intelligence and social cognition. 

While LLMs display ToM abilities and may well have access to orders of magnitude more accounts of internal mental states via the massive corpus of their pretraining data, none of that has been experienced first hand. Leaving aside that current LLMs likely have no mechanism for experience as we conceive of it \cite{chalmers2023large}, what are we to make of ToM in such alien minds?

\textbf{Game Theory}.
Our experiments and testing protocol assume users are unwise to model predictions about them. As users become aware that models are actively predicting their mental states and behavior, those predictions may become harder to make. Similarly, as LLMs take this into account, simulations will become still more complex.

\section{Discussion}
Principal-agent problems are a set of well understood coordination failures that emerge from interest misalignment and information asymmetry between persons or groups and their proxies. In normal political and economic life, delegating an agent incurs costs and efforts to minimize that risk reduce the efficiency of the agent.

We view our very early work in modeling user psychology as ultimately in service of eliminating the certitude of principal-agent problems from economic relations. As LLMs or other AI systems become increasingly capable and autonomous, they offer enormous economic potential. However, their alignment to human principals is not a foregone conclusion. On the contrary, we may instead see an \textit{exaggeration} of existing asymmetries between principals and agents, as well as the introduction of new concerns around latency, intelligence, and digital nativity.

In order to achieve trustworthy and efficient agentic AI, \textit{individual} alignment is required. Human agents and deterministic software are already capable of operating \textit{like} their principals. LLMs promise massive reductions in marginal cost along that axis, but hardly class better than the status quo (and often much worse) with regard to user alignment. Yet the unique potential here is agents \textit{who are} the principals themselves, that is, there is no meaningful practical or philosophical difference between discrete humans and the psychologically-aligned AIs acting on their behalf.

LLMs are excellent simulators capable of assuming myriad identities \cite{janus_2022}. They also excel at ToM tasks, and we've shown, can passively harvest and reason about user psychological data. These two interrelated qualities may very well make possible high-fidelity renderings of principals capable of flawlessly \textit{originating} and executing intent as their proxies with zero marginal agency cost. In this way LLMs may become more augmentation than tool, more appendage than agent.

\section{Acknowledgements}
The authors are grateful to Ayush Paul and Jacob Van Meter for their work on the Bloom development team, Thomas Howell of Forum Education for extensive conceptual review and ideation, and Zach Seward for invaluable advice and mentoring. We are additionally grateful to Ben Bowman for advising the machine learning aspects of this paper and Lee Ahern from the Bellisario College of Communications at Pennsylvania State University for feedback on the statistical tests and results section.

\printbibliography
\end{document}